\documentclass[sigconf]{acmart}

\usepackage{booktabs} 

\usepackage{algorithm}
\usepackage{algpseudocode}

\usepackage{caption}
\usepackage{subcaption}
\usepackage{graphicx}

\usepackage{balance}  

\setcopyright{rightsretained}

\copyrightyear{2018} 
\acmYear{2018} 
\setcopyright{acmlicensed}
\acmConference[KDD '18]{The 24th ACM SIGKDD International Conference on Knowledge Discovery \& Data Mining}{August 19--23, 2018}{London, United Kingdom}
\acmBooktitle{KDD '18: The 24th ACM SIGKDD International Conference on Knowledge Discovery \& Data Mining, August 19--23, 2018, London, United Kingdom}
\acmPrice{15.00}
\acmDOI{10.1145/3219819.3219837}
\acmISBN{978-1-4503-5552-0/18/08}

\begin{document}
\title{Autotune: A Derivative-free Optimization Framework for Hyperparameter Tuning}

\author{Patrick Koch}
\affiliation{\institution{SAS Institute Inc.}}
\email{Patrick.Koch@sas.com}

\author{Oleg Golovidov}
\affiliation{ \institution{SAS Institute Inc.}}
\email{Oleg.Golovidov@sas.com}

\author{Steven Gardner}
\affiliation{  \institution{SAS Institute Inc.}}
\email{Steven.Gardner@sas.com}

\author{Brett Wujek}
\affiliation{  \institution{SAS Institute Inc.}}
\email{Brett.Wujek@sas.com}

\author{Joshua Griffin}
\affiliation{  \institution{SAS Institute Inc.}}
\email{Joshua.Griffin@sas.com}

\author{Yan Xu}
\affiliation{ \institution{SAS Institute Inc.}}
\email{Yan.Xu@sas.com}

\renewcommand{\shortauthors}{P. Koch et al.}

\begin{abstract}
Machine learning applications often require hyperparameter tuning. 
The hyperparameters usually drive both the efficiency of the model training process and the resulting model quality. 
For hyperparameter tuning, machine learning algorithms are complex black-boxes. 
This creates a class of challenging optimization problems, whose objective functions tend to be nonsmooth, 
discontinuous, unpredictably varying in computational expense,
and include continuous, categorical, and/or integer variables. Further, function evaluations can fail 
for a variety of reasons including numerical difficulties or hardware failures.  
Additionally, not all hyperparameter value combinations are compatible, which creates so called hidden constraints. 
Robust and efficient optimization algorithms are needed for hyperparameter tuning. 
In this paper we present an automated parallel derivative-free optimization framework called \textbf{Autotune}, 
which combines a number of specialized sampling and search methods that
are very effective in tuning machine learning models despite these challenges. 
Autotune provides significantly improved models over using default hyperparameter settings with minimal user interaction 
on real-world applications. 
Given the inherent expense of training numerous candidate models, we demonstrate 
the effectiveness of Autotune's search methods and the efficient distributed and parallel paradigms for training and tuning models, 
and also discuss the resource trade-offs associated with the ability to both distribute the training process and 
parallelize the tuning process. 
\end{abstract}

%
%



\maketitle

\section{Introduction}
The approach to finding the ideal values for hyperparameters (tuning a model for
a particular data set) has traditionally been a manual effort. For guidance in
setting these values, researchers often rely on their past experience using
these machine learning algorithms to train models. However, even with expertise
in machine learning algorithms and their hyperparameters, the best settings of
these hyperparameters will change with different data; it is difficult to
prescribe the hyperparameter values based on previous experience. The ability to
explore alternative configurations in a more guided and automated manner is
needed.

A typical approach to generating alternative model configurations is through a
grid search. Each hyperparameter of interest is discretized into a desired set
of values to be studied, and models are trained and assessed for all
combinations of the values across all hyperparameters. Although easy to
implement, a grid search is quite costly because the computational expense grows exponentially
with the number of hyperparameters and the number of discrete levels of each.
While three hyperparameters
with three levels each requires only $27$ model configurations to be evaluated,
six hyperparameters with five levels each would require $15,625$ models to be
trained. Even with a substantial cluster of compute resources, training these many models
is prohibitive in most cases, especially with the computation
cost of modern machine learning algorithms and massive data sets associated with
applications like image recognition and natural language processing. 

A simple yet surprisingly effective alternative to performing a grid search is
to train and assess candidate models by using random combinations of
hyperparameter values. As demonstrated in Bergstra and Bengio\cite{BeBe12}, given the
disparity in the sensitivity of model accuracy to different hyperparameters, a
set of candidates that incorporates a larger number of trial values for each
hyperparameter will have a much greater chance of finding effective values for
each hyperparameter. Because some of the hyperparameters might actually have
little to no effect on the model for certain data sets, it is prudent to avoid
wasting the effort to evaluate all combinations, especially for
higher-dimensional hyperparameter spaces. Still, the
effectiveness of evaluating purely random combinations of hyperparameter values
is subject to the size and uniformity of the sample. Candidate combinations can
be concentrated in regions that completely omit the most effective combination
of values of the hyperparameters, and it is still likely to generate fewer
improved configurations. A recent variation on random search called Hyperband focuses on speeding up
random search by terminating ill-performing hyperparameter configurations (Li et
al.~\cite{LiJaDeRoTa17}).  This approach allows more configurations to be evaluated in a given time period,
increasing the opportunity to identify improved configurations. 

A approach similar to random search but more structured is to use a random Latin hypercube sample
(LHS) (McKay~\cite{Mc92}), an experimental design in which samples are exactly uniform
across each hyperparameter but random in combinations. These so-called
low-discrepancy point sets attempt to ensure that points are approximately
equidistant from one another in order to fill the space efficiently. This
sampling ensures coverage across the entire range of each hyperparameter and is
more likely to find good values of each hyperparameter which can then be used to
identify good combinations. Other experimental design procedures can also be
quite effective at ensuring equal density sampling throughout the entire
hyperparameter space, including optimal Latin hypercube sampling as proposed by
Sacks et al.~\cite{SaWeMiWy89}.

Exploring alternative model configurations by evaluating a discrete sample of
hyperparameter combinations, whether randomly chosen or through a more
structured experimental design approach, is certainly straightforward. However,
true optimization of hyperparameters should facilitate a complete search of
continuous parameter space in addition to discrete parameter space, and make use of
information from previously evaluated configurations to increase the number of
alternate configurations that show improvement. Discrete samples are unlikely to
identify even a local accuracy peak or error valley in the hyperparameter space;
searching between these discrete samples can uncover good combinations of
hyperparameter values. The search is based on an objective of minimizing the
model validation error, so each ``evaluation'' from the optimization algorithm's
perspective is a full cycle of model training and validation. Optimization
methods are designed to make intelligent use of fewer evaluations and thus save 
on the overall computation time. Optimization algorithms that have been used for
hyperparameter tuning include Broyden-Fletcher-Goldfarb-Shanno (BFGS) (Konen et
al.~\cite{KoKoFlBaFrNa11}), covariance matrix adaptation evolution strategy (CMA-ES) 
(Konen et al.~\cite{KoKoFlBaFrNa11}), particle swarm (PS) (Renukadevi and Thangaraj~\cite{ReTh14}; Gomes et al.
~\cite{GoPrSoRoCa12}), tabu search (TS) (Gomes et al.~\cite{GoPrSoRoCa12}), 
genetic algorithms (GA) (Lorena and
de Carvalho~\cite{LoCa08}), and more recently surrogate-based Bayesian optimization
(Denwancker et al.~\cite{DeMcClHaJoKe16}). 

\begin{figure}
\includegraphics[width=.99\linewidth]{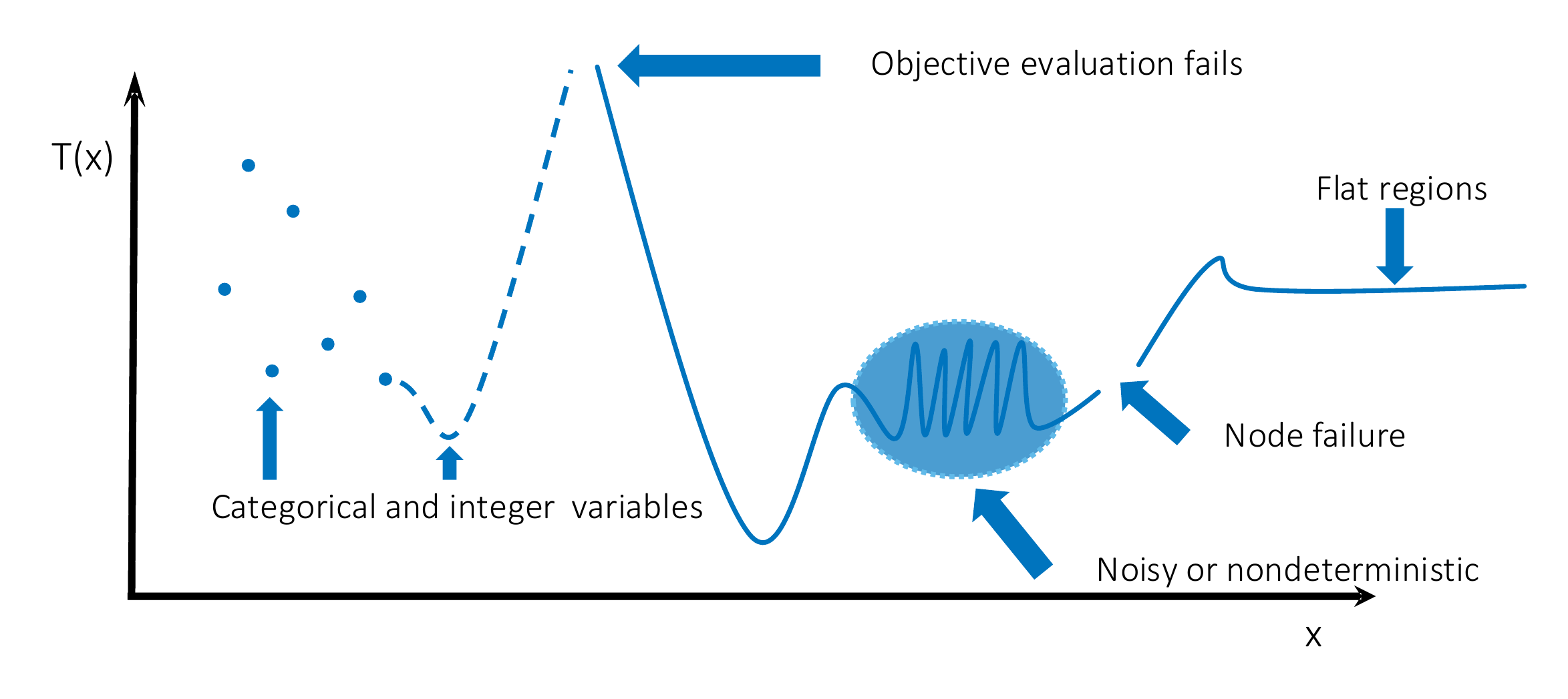}
\caption{Challenges in applying optimization to hyperparameter tuning.}
\label{fig1}
\end{figure}

However, because machine learning training and scoring algorithms are a complex
black-box to the tuning algorithm, they create a class of challenging
optimization problems. Note that optimization variables are hyperparameters here. 
Figure \ref{fig1} illustrates several of these challenges: 
\begin{itemize}
\item   Machine learning algorithms typically include not only continuous
variables, but also categorical and integer variables, leading to a very discrete
objective space.
\item   In some cases, the variable space is discontinuous, and the
objective evaluation fails.
\item   The space can also be very noisy and nondeterministic, for example, when
distributed data are moved around because of unexpected rebalancing.
\item   Objective evaluations can fail because of numerical difficulties or hardware failures, which can
derail a search process.
\item   Often the search space contains many flat regions where multiple configurations
produce very similar models and an optimizer can fail to find a direction of
improvement.
\end{itemize}

An additional challenge is the unpredictable computational expense of training and
validating predictive models using different hyperparameter values.  Adding
hidden layers and neurons to a neural network, for example, can significantly
increase the training and validation time, resulting in widely ranging potential
objective expense. Given these challenges, a very flexible and efficient search
strategy is needed.  As with machine learning algorithms, the no free lunch
theorem applies to optimization algorithms (Wolpert~\cite{Wo96}; Wolpert and Macready~\cite{WoMa97}), 
i.e., no single algorithm can overcome all these challenges and work well for all
data sets.  Also, the strengths of sampling methods cannot be overlooked.

In the next section, we introduce our automated parallel derivative-free
optimization framework \textbf{Autotune} that concurrently exploits the strengths of sampling
methods and multiple derivative-free optimization algorithms,
which are very effective for hyperparameter tuning.  Given the inherent expense of
training numerous candidate models, we then discuss efficient distributed and
parallel paradigms for training and tuning models, and also discuss the resource
tradeoffs associated with the ability to both distribute the training process
and parallelize the tuning process. Finally, we report benchmark tuning results,
present two case studies, and conclude with contributions and future work. 


\section{Derivative-free OPTIMIZATION FRAMEWORK}

In this section, we describe the derivative-free optimization framework Autotune, the search
methods incorporated, and its default search method.
Autotune is a product within SAS\textsuperscript{\textregistered} ~Visual Data Mining and Machine Learning \cite{WeHaMy17}, and
operates on SAS\textsuperscript{\textregistered} ~Viya\textsuperscript{\textregistered} \cite{Viya}, which is designed to enable 
distributed analytics and to support cloud computing.
Autotune is able to tune the hyperparameters of various machine learning models including
decision trees, forests, gradient boosted trees, neural networks, support vector machines, 
factorization machines, and Bayesian network classifiers. 

\subsection{System Overview} 

Autotune is designed to perform optimization of
general nonlinear functions over both continuous and integer variables. The
functions do not need to be expressed in analytic closed form, black-box
integration is supported, and they can be non-smooth, discontinuous, and
computationally expensive to evaluate. Problem types can be single-objective or
multiobjective.  The system is designed to run in either single-machine mode or
distributed mode.

Because of the limited assumptions that are made about the objective function 
and constraint functions, Autotune takes a parallel hybrid derivative-free
approach similar to those used in Taddy et al.~\cite{TaGr09}; Plantenga~\cite{Pl09}; Gray,
Fowler, and Griffin~\cite{GrFoGr10}; Griffin and Kolda~\cite{GrKo10}. Derivative-free methods
are effective whether or not derivatives are available, provided that the
dimension of x is not too large (Gray and Fowler~\cite{GrFo11}). As a rule of thumb,
derivative-free algorithms are rarely applied to black-box optimization problems
that have more than 100 variables. The term ``black-box'' emphasizes that the
function is used only as a mapping operator and makes no implicit assumption
about or requirement on the structure of the functions themselves. In contrast,
derivative-based algorithms commonly require the nonlinear objectives and
constraints to be continuous and smooth and to have an exploitable analytic
representation.

Autotune has the ability to simultaneously apply multiple
instances of global and local search algorithms in parallel. This streamlines
the process of needing to first apply a global algorithm in order to determine a
good starting point to initialize a local algorithm. For example, if the problem
is convex, a local algorithm should be sufficient, and the application of the
global algorithm would create unnecessary overhead. If the problem instead has
many local minima, failing to run a global search algorithm first could result
in an inferior solution. Rather than attempting to guess which paradigm is best,
the system simultaneously performs global and local searches while continuously
sharing computational resources and function evaluations. The resulting run time
and solution quality should be similar to having automatically selected the best
global and local search combination, given a suitable number of threads and
processors. Moreover, because information is shared among simultaneous searches, the robustness of this
hybrid approach can be increased over other hybrid combinations that simply use the
output of one algorithm to hot-start the second algorithm. 

Inside Autotune, integer and categorical variables are handled by using strategies and
concepts similar to those in Griffin et al.~\cite{GrFoGrHe11}. This approach can be viewed
as a genetic algorithm that includes an additional ``growth'' step, in which
selected points from the population are allotted a small fraction of the total
evaluation budget to improve their fitness score (that is, the objective
function value) by using local optimization over the continuous variables.

This Autotune framework supports:
\begin{itemize}
\item Running in distributed mode on a cluster of machines that distribute the data and the computations
\item Running in single-machine mode on a server 
\item Exploiting all the available cores and concurrent threads, regardless of execution mode
\end{itemize}

\begin{figure}
\includegraphics[width=.99\linewidth]{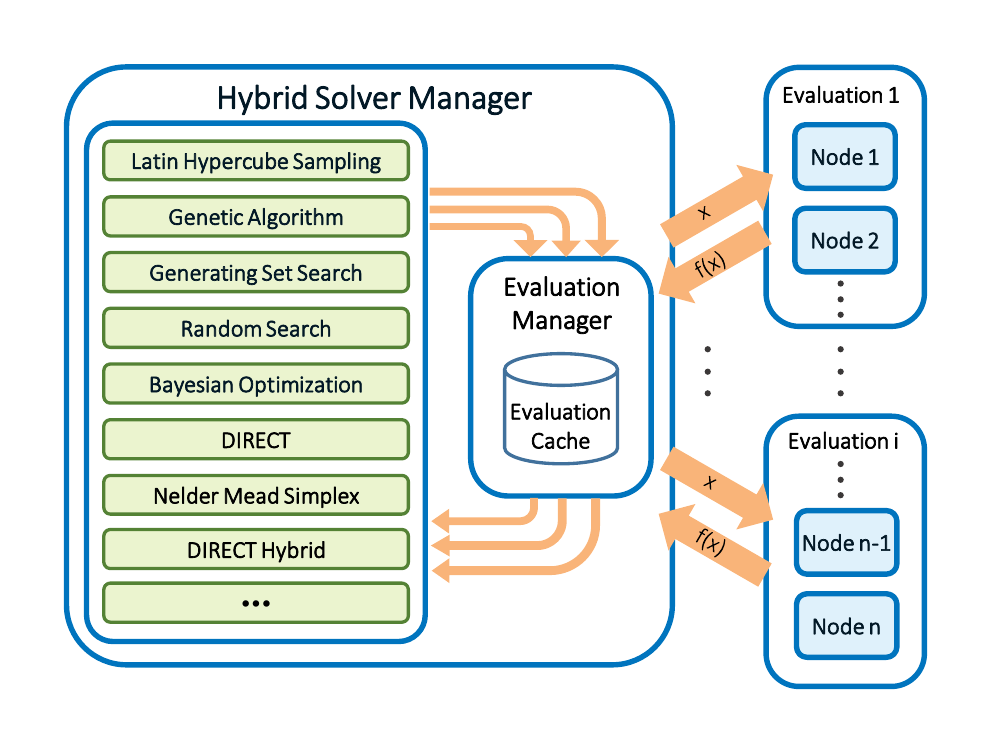}
\caption{The Autotune framework.}
\label{fig2}
\end{figure}

A pictorial illustration of this framework is shown in Figure \ref{fig2}. An extendable suite of
search methods (also called solvers) are driven by the Hybrid Solver Manager that controls concurrent
execution of the search methods. New search methods can easily be added to the framework. 
Objective evaluations are distributed across
multiple worker nodes in a compute grid and coordinated in a feedback loop that
supplies data from running search methods. 

Execution of the system is iterative in its processing, with each iteration
containing the following steps:
\begin{enumerate}
\item Acquire new points from the solvers
\item   Evaluate each of those points by calling the appropriate black-box functions (model training and validation)
\item   Return the evaluated point values (model accuracy) back to the solvers
\item   Repeat
\end{enumerate}

For each solver in the list, the evaluation manager exchanges points with that
solver.  During this exchange, the solver receives back all the points that were
evaluated in the previous iteration.  Based upon those evaluated point values,
the solver generates a new set of points it wants evaluated and those new points
get passed to the evaluation manager to be submitted for evaluation.  For
any solvers capable of ``cheating'', they may look at evaluated points that were
submitted by a different solver.  As a result, search methods can learn from
each other, discover new opportunities, and increase the overall robustness of
the system.

\subsection{Search Methods}

Autotune is designed to support multiple search methods, 
which not only can be run concurrently but they also can be combined to
create new hybrid methods.  In addition to the sampling methods (random and LHS)  already
discussed and the default search method to be introduced in the next session, 
the set of supported search methods include the following:

\subsubsection{Genetic Algorithm (GA)} 
GAs are a family of search algorithms
that seek optimal solutions to problems by applying the principles of natural
selection and evolution (Goldberg~\cite{Go89}). Genetic algorithms can be applied to
almost any optimization problem and are especially useful for problems for which
other calculus-based techniques do not work, such as when the objective function
has many local optima, when the objective function is not differentiable or
continuous, or when solution elements are constrained to be integers or
sequences. In most cases, genetic algorithms require more computation than
specialized techniques that take advantage of specific problem structures or
characteristics. However, for optimization problems for which no such techniques
are available, genetic algorithms provide a robust general method of solution. 

\subsubsection{Generating Set Search (GSS)} This type of method is designed for problems that have
continuous variables and have the advantage that, in practice, they often
require significantly fewer evaluations to converge than an exploratory search
method like GA (Griffin and Kolda~\cite{GrKo10a}). GSS can provide a measure of local
optimality that is very useful in performing multimodal optimization. It may add
additional ``growth steps'' to an exploratory search method for continuous
variables. 

\subsubsection{Bayesian Optimization}
The Bayesian optimization method in Autotune employs a Gaussian process surrogate model ~\cite{Jo01}.  
LHS is used to initialize the surrogate model, which is then used to generate new
evaluations that minimize the approximate function.  These new evaluations are
executed using the real black-box function and potentially added to the surrogate
model for increased accuracy until a certain maximum number of points are in the approximate model.
Confidence levels between samples and an exploration parameter allows generations of trials in new regions to avoid
converging on lower accuracy models.

\subsubsection{DIRECT}
This method is an implicit branch and bound type
algorithm that divides the hyper-rectangle defined by the variable bounds into
progressively smaller rectangles where the relevance of a given rectangle is
based on its diameter and the objective value at the center point ~\cite{JoPeSt93}.  The former
is used to quantify uncertainty, the latter is used to estimate the best value
within.  A Pareto set is maintained for these two quantities and used to select
which of the hyper-rectangles to trisect at the next iteration. 

\subsubsection{Nelder-Mead}
This method is a variable shape simplex
direct-search optimization method that maintains the objective values of the
vertices of a polytope whose number is one greater than the dimension being
optimized~\cite{NeMe65}.  It then predicts new promising vertices for the simplex based on
current values using a variety of simplex transformation operations. 

\subsubsection{DIRECT Hybrid}
This hybrid method first uses
DIRECT principles to divide and sort the feasible regions into a set of
hyper-rectangles of varying dimension based on the likelihood of containing a global
minimizer.  As the hyper-rectangles are divided, the size of the rectangles as
measured by the distance between its center and corners reduces.  When this size
is small enough, then a Nelder-Mead optimization is executed based on the small
hyper-rectangle to further refine the search and the small hyper-rectangle is no
longer considered for division.  The best value found by a small
hyper-rectangle's Nelder-Mead optimizer is then used to represent that given rectangle.

\subsection{Default Search Method}

\begin{figure}
\includegraphics[width=.99\linewidth]{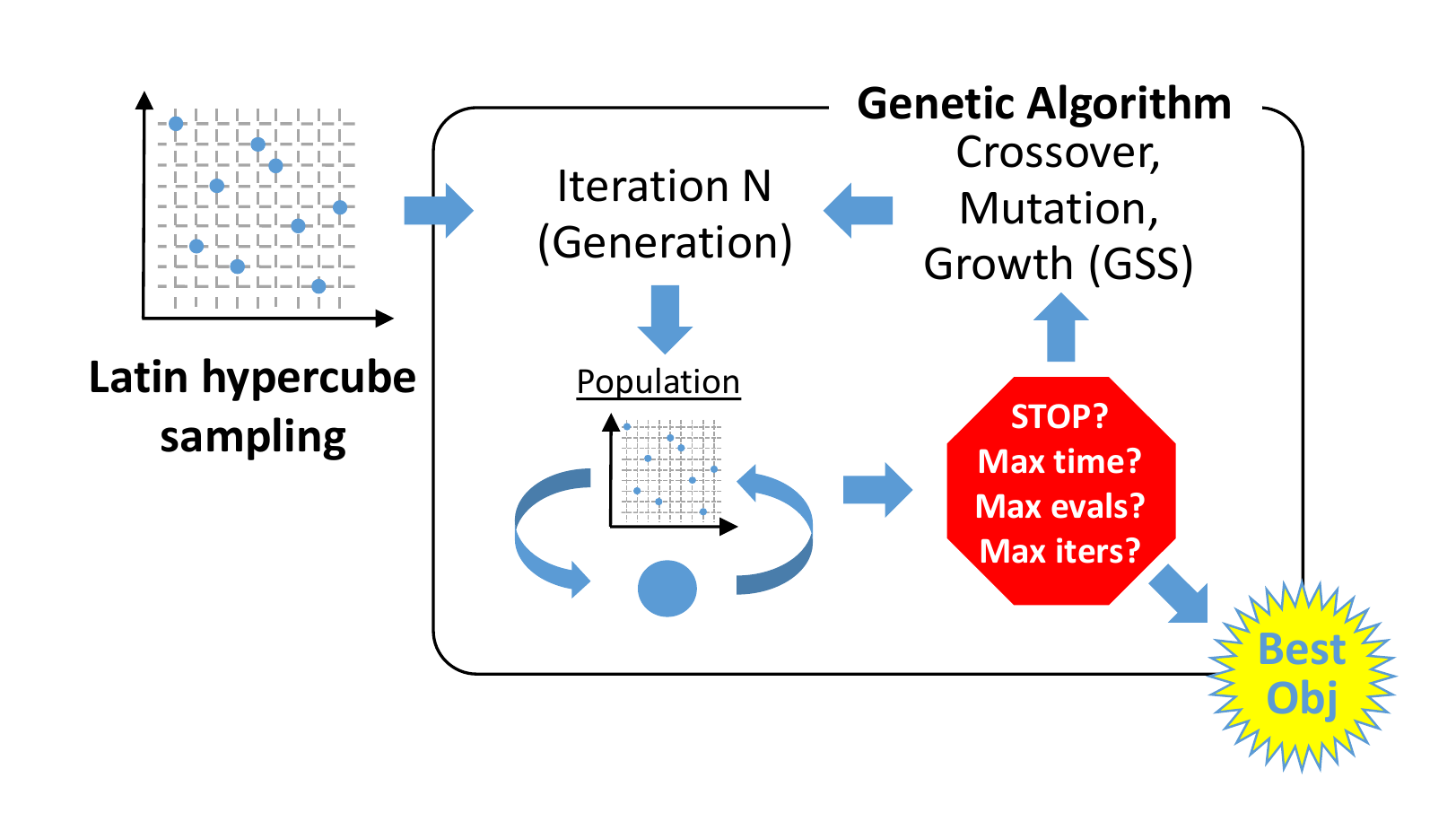}
\caption{The tuning process of the default search method used by Autotune.}
\label{fig3}
\end{figure}
 
\begin{algorithm}
\caption{Default Search Method in Autotune}
  \label{alg:outeralg}
  \begin{algorithmic}[1]
    \Require Population size $n_p$, and evaluation budget $n_b$.
    \Require Number of centers $n_c < n_p$ and initial step-size $\hat \Delta$.
    \Require Sufficient decrease criterion $\alpha \in (0,1)$. 
    \State Generate initial parent-points ${\mathcal P} $ using LHS with $|{\mathcal P}| = n_p$. 
    \State Evaluate ${\mathcal P}$ asynchronously in parallel.
    \State Populate reference cache-tree, ${\mathcal R},$ with unique points from ${\mathcal P}$.
    \State Associate each point $p \in {\mathcal P}$ with step $\Delta_p$ initialized to $\hat \Delta$.
    \While{$(|{\mathcal R}| \le n_b)$}
    \State Select ${\mathcal A} \subset {\mathcal P}$ for local search, such that $|{\mathcal A}| = n_c$.
        \For {$p \in {\mathcal A}$} \Comment{Search along compass directions}
         \State Set ${\mathcal T}_p = \{ \}$
        	\For {$e_i \in I$}
        		\State ${\mathcal T}_p = {\mathcal T}_p \cup \{p+\Delta_p e_i\}\cup \{p-\Delta_p e_i\}$
        	\EndFor 
    \EndFor
    \State Generate child-points ${\mathcal C}$ via crossover and mutations on ${\mathcal P}$.
    \State Set ${\mathcal T} = {\mathcal C} \cup_{p \in {\mathcal A}}{\mathcal T}_p$. 
    \State Evaluate ${\mathcal T} \cap {\mathcal R}$ using fast tree-search look-up on ${\mathcal R}$.
    \State Evaluate remaining ${\mathcal T} - {\mathcal R}$ asynchronously in parallel.
    \State Add unique points from  ${\mathcal T} - {\mathcal R}$ to cache-tree ${\mathcal R}$. 
    \State Update ${\mathcal P}$ with new generation ${\mathcal C}$ and initial step $\hat \Delta$.

    \For{$p \in {\mathcal A}$} 
    \If{$\min_{ y \in {\mathcal T}_p} f(y) < f(p) - \alpha \Delta_{p}^2$}
    \State  Set $p = y$ \Comment{Pattern search success}
    \Else
    \State Set $\Delta_p = \Delta_p/2$ \Comment{Pattern search failure}
    \EndIf
    \EndFor
    \EndWhile
  \end{algorithmic}
\end{algorithm} 

As illustrated in Figure \ref{fig3} and explained by the pseudocode in Algorithm  \ref{alg:outeralg}, 
the default search method used by Autotune is a
\emph{hybrid} method that begins with a Latin hypercube sample of the
hyperparameter space.  The best configurations from the LHS are then used to generate the initial population
for the GA, which crosses and mutates the best samples in an iterative process
to generate a new population of model configurations at each iteration.  

In addition to the crossover and mutation operations of a classic GA, Autotune
adds an additional "growth" step to each iteration of the GA.  This permits the GSS algorithm
to perform local search in a neighborhood of select members from the current GA population.  
This can improve convergence to a good minimum once the GA is sufficiently near the
corresponding basin or region of attraction.  Typically the best point in the GA population
is continuously optimized.  If sufficient computing resources are available, 
other points may be optimized simultaneously by, for example, selecting points randomly 
from the Pareto-front comparing the population's objective function and distance to the nearest neighbor.

The default search method in Autotune essentially combines the elements of LHS, GA and GSS methods. 
The strengths of this hybrid method include handling of continuous, integer,
and categorical variables; handling nonsmooth, discontinuous spaces; and ease of
parallelizing the search. All are prevalent and critical for the
hyperparameter tuning problem.

Autotune uses a specified model accuracy measure (misclassification,
mean squared error, multiclass log loss, AUC, KS coefficient, etc.) as objective values.  
This measure is calculated on validation data,  otherwise the autotuning process would likely
overfit the training data.  Validation is an additional, but necessary, expense
during tuning when training many alternative model configurations.  Ideally a
cross-validation process is applied to incorporate all data in training and
validation, with separate ``folds''.  However, evaluation of each fold for each
model configuration significantly increases the training expense and thus the
tuning expense, making it prohibitive for big data applications.  Fortunately, 
it is often unnecessary and undesirable to run each training process to
completion when tuning. Given information about the current best configurations, it is
possible to abort running model configurations after a subset of all folds if
the estimated model quality is not near the current best.  This is one form of
early stopping that is supported by Autotune.

Even with aborting of bad models, many datasets are still too large for
cross-validation.  In this case, a single validation partition is used. 
To ensure that the training subset and the validation subset are both representative of the original data set,
stratified sampling is used when possible (nominal target) .  
With very large data sets, subsampling can also be employed
to reduce the training and validation time during tuning, and again stratified
sampling helps ensure the data partitions remain representative.   
The biggest increase in efficiency, however, comes from the evaluation of alternate model
configurations in parallel - a process that comes with its own set of challenges.  
The parallel hyperparameter tuning implementation in Autotune is detailed in the next section.


\section{PARALLEL HYPEPARAMETER TUNING}\label{section3}

The training of a model by a machine learning algorithm is often computationally
expensive. As the size of a training data set grows, not only does the expense
increase, but the data (and thus the training process) must often be distributed
among compute nodes because the data exceed the capacity of a single computer.
Also, the configurations to be considered during tuning are independent, making
a sequential tuning process not only expensive but unnecessary, given a grid of
compute resources. 

While some systems only support assigning worker nodes to either the training
process or the tuning process (sequential tuning with each model trained on all
workers or parallel training of multiple models each on one worker), the
Autotune system presented here supports both assigning multiple worker nodes to
each model to be trained and training multiple models in parallel. The challenge
is to determine the best usage of available worker nodes for the tuning process as a whole.

\begin{figure}
\begin{subfigure}{.25\textwidth}
  \centering
  \includegraphics[width=.9\linewidth]{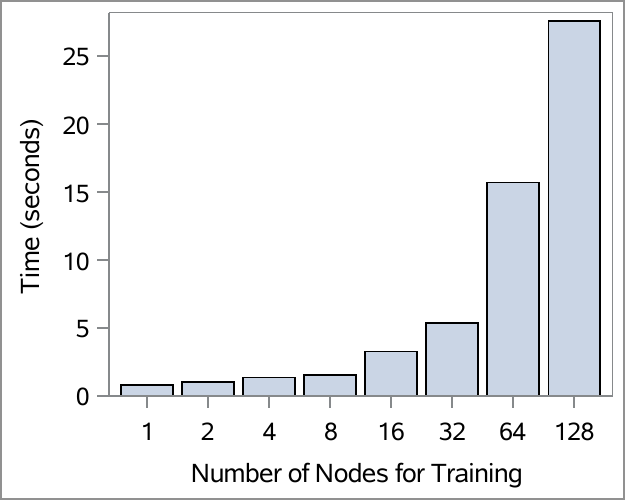}
  \caption{150 obs, 5 columns}
  \label{fig4:sub150}
\end{subfigure}%
\begin{subfigure}{.25\textwidth}
  \centering
  \includegraphics[width=.9\linewidth]{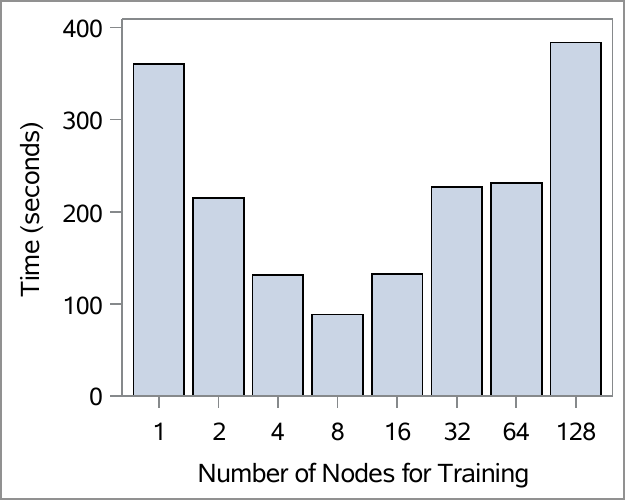}
  \caption{581k obs, 54 columns}
  \label{fig4:sub581k}
\end{subfigure}
\caption{Training times with different number of computing nodes on two data sets.}
\label{fig4}
\end{figure}

For small data sets data distribution is not necessary, but it may not be clear
that it can actually be detrimental, reducing performance.  In Figure ~\ref{fig4:sub150}, a
tree based gradient boosting algorithm is applied to train a model to the
popular iris data set (containing only 150 observations) using a number of
different worker nodes ranging from 1 to 128.  The communication cost required to
coordinate data distribution and model training increases continuously as the number of worker nodes
increases. The training time grows from less than 1 second on a single machine to nearly half a
minute on 128 nodes. In this case,  a model tuning process would benefit 
more from parallel tuning (training different model configurations in parallel) 
than from distributed/parallel training of each model; with a grid of 128 nodes, 128 models could be trained in parallel
without overloading the grid.

As shown in Figure ~\ref{fig4:sub581k}, 
for larger data sets, distributing the data and the training process reduces the
training time; here gradient boosting and covertype \footnote{The Forest Covertype dataset is Copyrighted
1998 by Jock A. Blackard and Colorado State University.} data set are used. 
The covertype data set contains over 581K observations and 54 features.  
However, the benefit of data distribution and parallel training does not continue to increase
with an increasing number of worker nodes. At some point the cost of
communication again outweighs the benefit of parallel processing for model
training. Here the time for training increases beyond 8 worker nodes, to a point where
32 and 64 nodes are more costly than 2  nodes and using all 128 is more
costly than using only 1 node. 

Determining the best worker node allocation for the tuning process is more
challenging than determining the most efficient training process.  
In Figure ~\ref{fig4:sub581k}, the training process is most efficient with 8 worker nodes.  
However, a grid of 128 nodes would support 16 different model configurations trained in
parallel during tuning if each uses 8 worker nodes (without overloading the
grid).  The training expense is not half with 8 worker nodes compared to with 4
worker nodes, and so it may make more sense to train each model with 4 worker
nodes, allowing 32 model configurations to be trained in parallel.  In fact, if
the data fits on one worker node, 128 model configurations trained in parallel
on 1 worker each may be more efficient than 4 batches of 32 models each trained
on 4 workers.  For very large data sets, the data must be distributed, but
training multiple models in parallel typically leads to larger gains in tuning
efficiency than training each model faster by using more worker nodes for each
model configuration. The performance gain becomes nearly linear as the number of
nodes increases because each trained model is independent during tuning, so no
communication is required between the different configurations being trained.
Determining the right resource allocation then depends on the size of the data
set, the size of the compute grid, and the tuning method taken. Note that an iterative
search strategy limits the size of each parallel batch (for example, the population size at
each iteration of the genetic algorithm).  

\begin{figure}
\includegraphics[width=.65\linewidth]{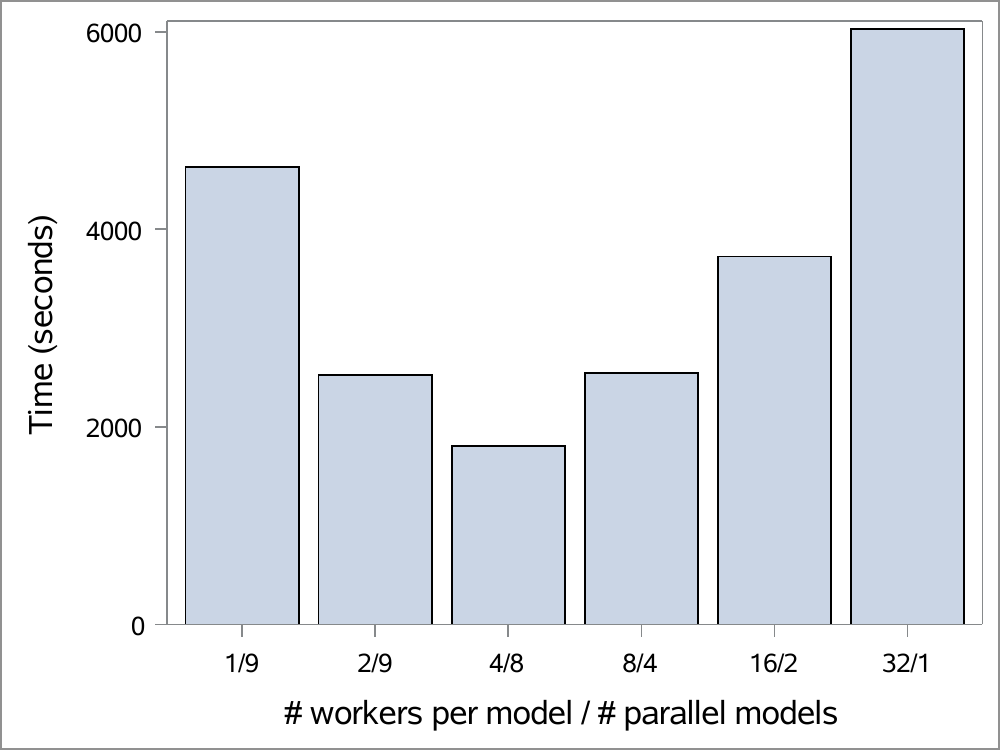}
\caption{A tuning resource allocation example.} 
\label{fig5}
\end{figure}

Allocating resources to both the model training process and the model tuning
process requires very careful management of the data, the training process, and
the tuning process.  Multiple alternate model configurations are submitted
concurrently by the framework, and the individual
model configurations are trained and validated on a subset of available worker
nodes in isolated processes. This allows multiple nodes to be used to manage large
training data when necessary, speeding up each individual training process. Figure ~\ref{fig5} shows 
a tuning time comparison for tuning the gradient boosting model to the covertype
data set. The tuning process consists of 5 iterations of 10 models and uses 
the default search method on a compute grid of  32 workers.
In this case, 4 workers for each model configuration, with 8 parallel
configurations is most efficient.  However, up to 16 configurations could be
evaluated in parallel with 2 workers rather than 8 with 4 workers, nearly
doubling the total number of configurations over 5 iterations without doubling
the tuning time.

\begin{figure}
\begin{subfigure}{.5\textwidth}
  \centering
  \includegraphics[width=.99\linewidth]{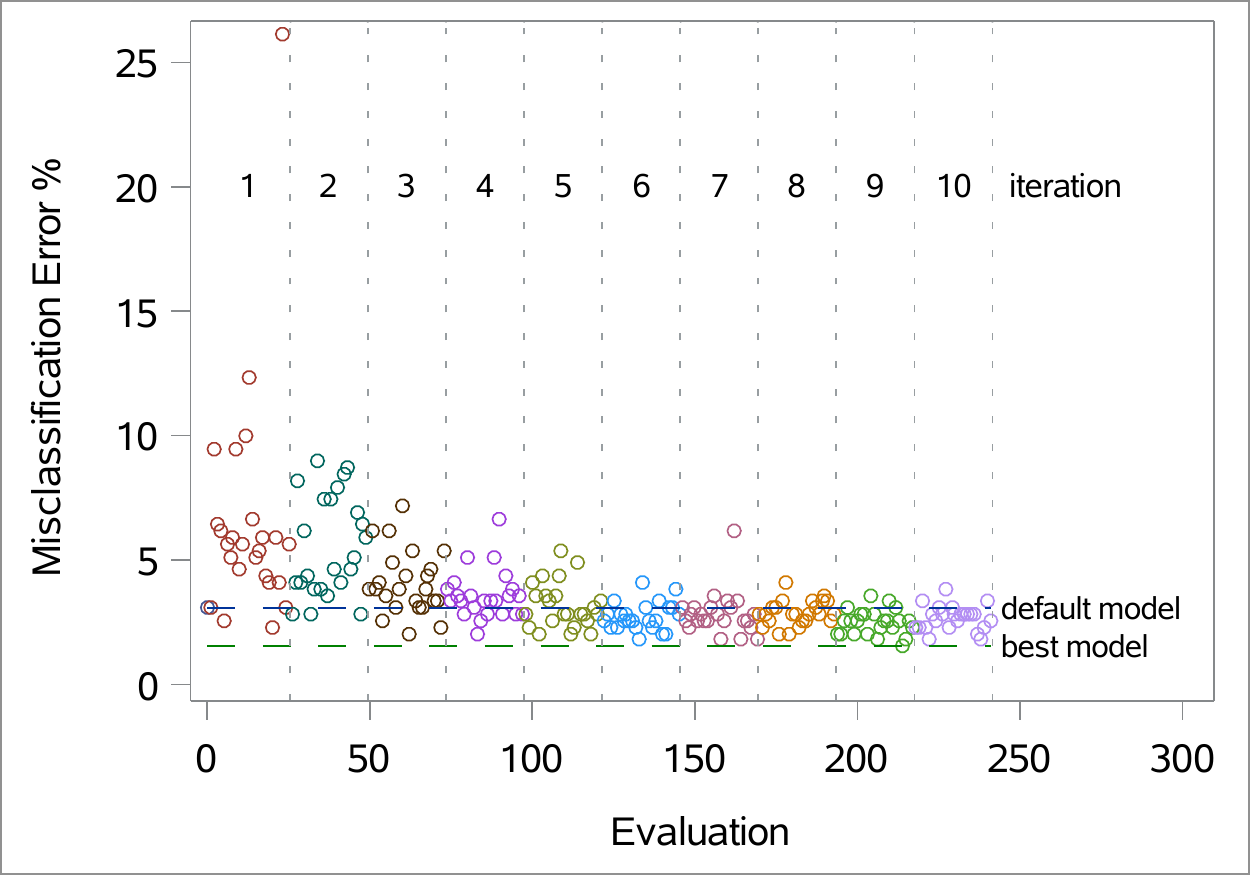}
  \caption{The default hybrid search history}
  \label{fig6:GA}
\end{subfigure}%
\hfill
\begin{subfigure}{.5\textwidth}
  \centering
  \includegraphics[width=.99\linewidth]{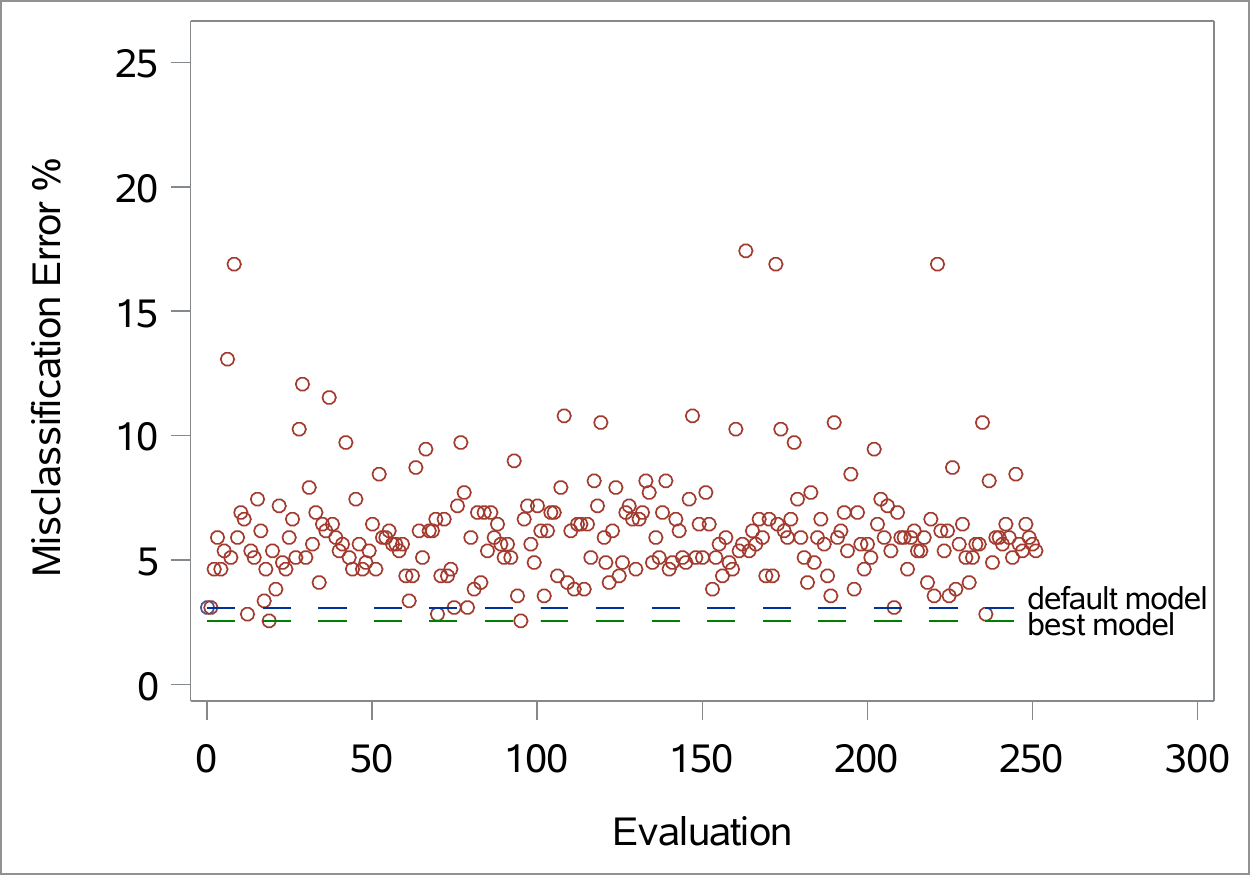}
  \caption{Random search history}
  \label{fig6:random}
              \begin{minipage}{.1cm}
            \vfill
            \end{minipage}
\end{subfigure}
\caption{The tuning history for the default hybrid search and random search on an image recognition data set.}
\label{fig6}
\end{figure}

\begin{figure*}
\includegraphics[width=.99\linewidth]{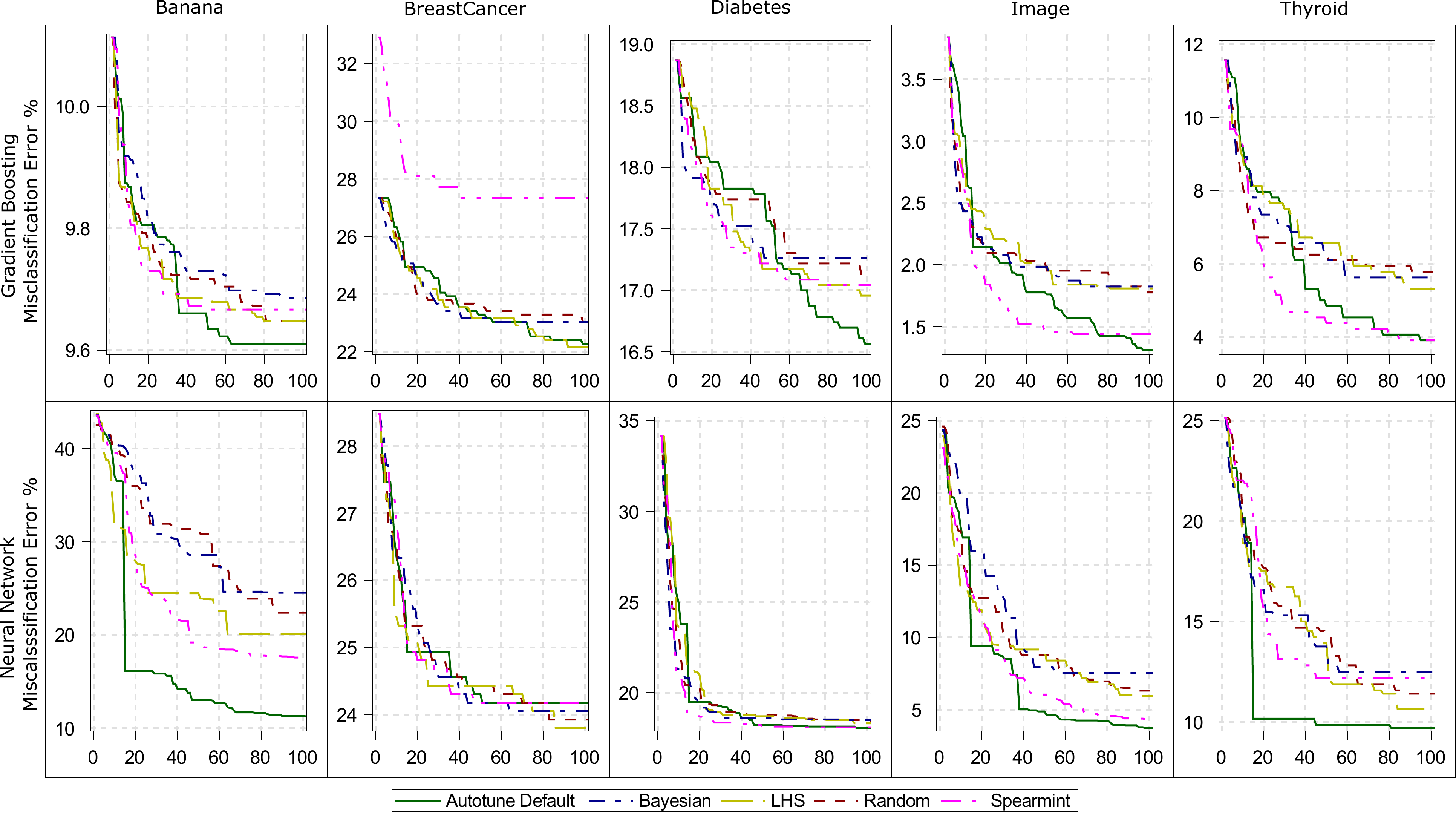}
\caption{Benchmark experiment results.} 
\label{fig7}
\end{figure*}

When it comes to choosing a search method for automated parallel
hyperparameter tuning, time, available compute resources, and tuning goals drive
the choice. Random search is popular for two main reasons: a) the hyperparameter
space is often discrete, which random search naturally accommodates, and b) random
search is simple to implement and all hyperparameter configurations could potentially be
evaluated concurrently because they are all independent and can be
pre-specified. The latter reason is a strong argument when a limited number of
configurations is considered or a very large compute grid is available. Figure ~\ref{fig6}
illustrates the tuning history of Autotune's default search method and
random search using an image recognition data set.  Here 10 iterations of 25
configurations are performed with the default hybrid approach and a single sample of
250 configurations for random search.  The learning occurring through the
optimization strategy  can clearly be seen in Figure  ~\ref{fig6:GA}. The initial
iteration contains configurations most of which are worse than the
initial/default, but as the iterations progress, more and more improvements are
found with the last iteration containing mostly improved configurations.  In the
case of random sampling, the results are fairly uniform across the history of
250 configurations chosen, as expected; however, many fewer improvements are
identified.  If the final ``best'' models are similar and 250 grid nodes are
available, and the data can fit on one worker node, the random search will be
more efficient.  However, if less than 250 grid nodes are available and/or if a
comparison and selection among top improved models is sought, the hybrid
search method that learns the more effective configurations across multiple iterations is more effective.



\section{Experiments} \label{section4}

To evaluate the performance of Autotune and the effectiveness of each search method,
 we conducted a benchmark experiment by applying the Autotune system to 
a set of five familiar benchmark data sets. 
The five data sets are taken from mldata.org ~\cite{mldata}, and include
\textit{banana}, \textit{breast cancer}, \textit{diabetes}, \textit{image} and \textit{thyroid}. 
All problems are tuned with a $30\%$ single partition for error validation during tuning.
For the default hybrid search method and Bayesian search method, 
10 iterations with 10 hyperparameter configurations per iteration are used; 
for random search and LHS, the sample size is 100.
All problems are run 10 times, and the results obtained are averaged to 
better assess behavior of the search methods. We also use 
the open source Spearmint Bayesian optimization package \cite{Snoek} for comparison.  

Two model types are used in this experiment. 
For tree-based gradient boost models, six hyperparameters are tuned:  
\textit{number of trees}, \textit{number of inputs to try when splitting}, \textit{learning rate}, \textit{sampling rate}, 
\textit{lasso}, and \textit{ridge regularization}. For fully connected neural network models, seven hyperparameters are tuned:
 \textit{number of hidden layers (0-2)}, \textit{number of neurons in each hidden layer}, 
 \textit{L1 and L2 regularization}, \textit{learning rate}, and \textit{annealing rate}.  
 
Results for tuning both model types are shown in Figure ~\ref{fig7}.  
For tuning the gradient boosting models, the default method performs better on three of the five data sets;
 LHS or Spearmint is each slightly better than the default on one data set. 
For tuning the neural network model, again, the default method performs better on three of the five data sets, and
LHS or Spearmint each wins one. 
These results show that the default method used by Autotune is very competitive and robust, and
an effective hyperparameter tuning system needs to employ a suite of 
diversified search methods to cover a wide range of problems. Furthermore,
integrating/combining different search methods is an effective way to create powerful hybrid methods.


\section{Case Studies}

Autotune has been deployed in many real-world applications. Here we report 
the use of Autotune to find better models in two applications. 


\subsection{Bank Product Promotion Campaign}

The bank data set ~\cite{bankdata} consists of anonymized and transformed observations taken from 
a large financial services firm's accounts, and contains 1,060,038 observations and 21 features. 
Accounts in the data represent
attributes describing the customer's propensity to buy products, RFM (recency, frequency, 
and monetary value) of previous transactions, and characteristics related to profitability
and creditworthiness. 
The goal is to predict which customers to target as the most likely to purchase new bank products 
in a promotional campaign.

The compute grid available for this study contains 40 machines:  a controller node and 39 worker nodes.  
Each model train uses 2 worked nodes, which allows 19 hyperparameter configurations to be evaluated in parallel without overloading the grid. 

In this study, we investigate the convergence properties of different search methods. 
Due to the long running time of Spearmint Bayesian method on large data sets, it is not included our case studies.
The default search method is configured with a population size of 115 
(resulting in 6 batches of 19 plus the default/best configuration).  The number of iterations 
is set to 20, resulting in up to 2281 model configurations evaluated. 
Random search and LHS are set to the same total sample size of 2280 plus default.  
The Bayesian search method is configured to run 60 iterations of 38, allowing model updating after two parallel batches of 19, 
with a matching maximum number of evaluations of 2280. 
Each search method is executed 10 times to average random effects.
We use a tree-based gradient boosting model and tune six of its hyperparameters as listed in Section \ref{section4} . 
For each search method, the tuning  takes from 1 to 3 hours, so it takes roughly 1 full day to run each 10 times.

\begin{figure}
  \centering
  \includegraphics[width=.99\linewidth]{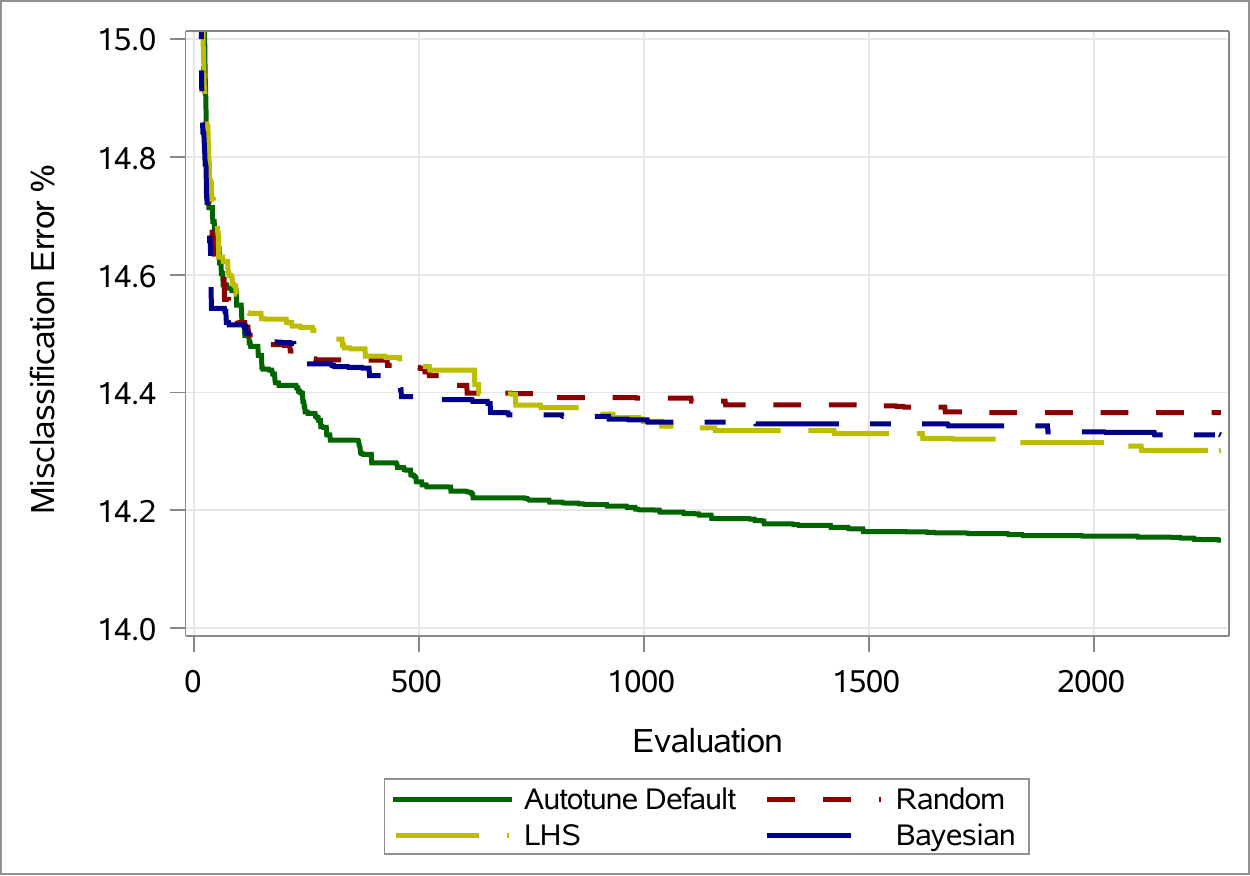}
              \begin{minipage}{.1cm}
            \vfill
            \end{minipage}
\caption{The autotune results of the bank product promotion data set. The gradient boost model error 
              with default hyperparameter settings is nearly $20\%$. For clarity of convergence comparisons \
              only the last percent of improvement, below $15\%$, is shown here.}
\label{fig8}
\end{figure}

Tuning results for the bank data are shown in Figure  ~\ref{fig8}.
 It is clear that all the tested methods are able to find better hyperparameter configurations quickly.
The errors are reduced from $20\%$  to  $15\%$ during the first few batches of evaluations. 
 The last percent of improvement happens gradually 
over the remaining $2200+$ evaluations, at different rates and with different final `best' results 
for each search method. 
After around 100 evaluations, the Bayesian search, random search, and LHS begin to stagnate 
while the default method continues to learn and reduce the model error and outperforms the other methods. 

It is important to note here that the `2X random' approach that has also 
become a popular basis for comparison is not relevant in this case.  
Since we can only run 19 models in parallel, and are doing so for all search methods, 
2X random will not be more efficient.  It may or may not find equivalent or better solutions,
but will take twice as long given this grid configuration.  Academically, the argument is valid: 
if we had 4000 machines, running 2000 configurations in parallel with 2 worker nodes each 
would be the most efficient.  Realistically, most data scientists do not have access 
to that many resources, and must share the resources that are available.  
Also, for this study, 1000 evaluations used in the default search method 
result in a better model than 2000 random samples; 
even at this level of resource allocation, the intelligent search methods are able 
to find improvement beyond those found by twice as many random samples.


\subsection{Wine Quality}

\begin{figure}
\includegraphics[width=.99\linewidth]{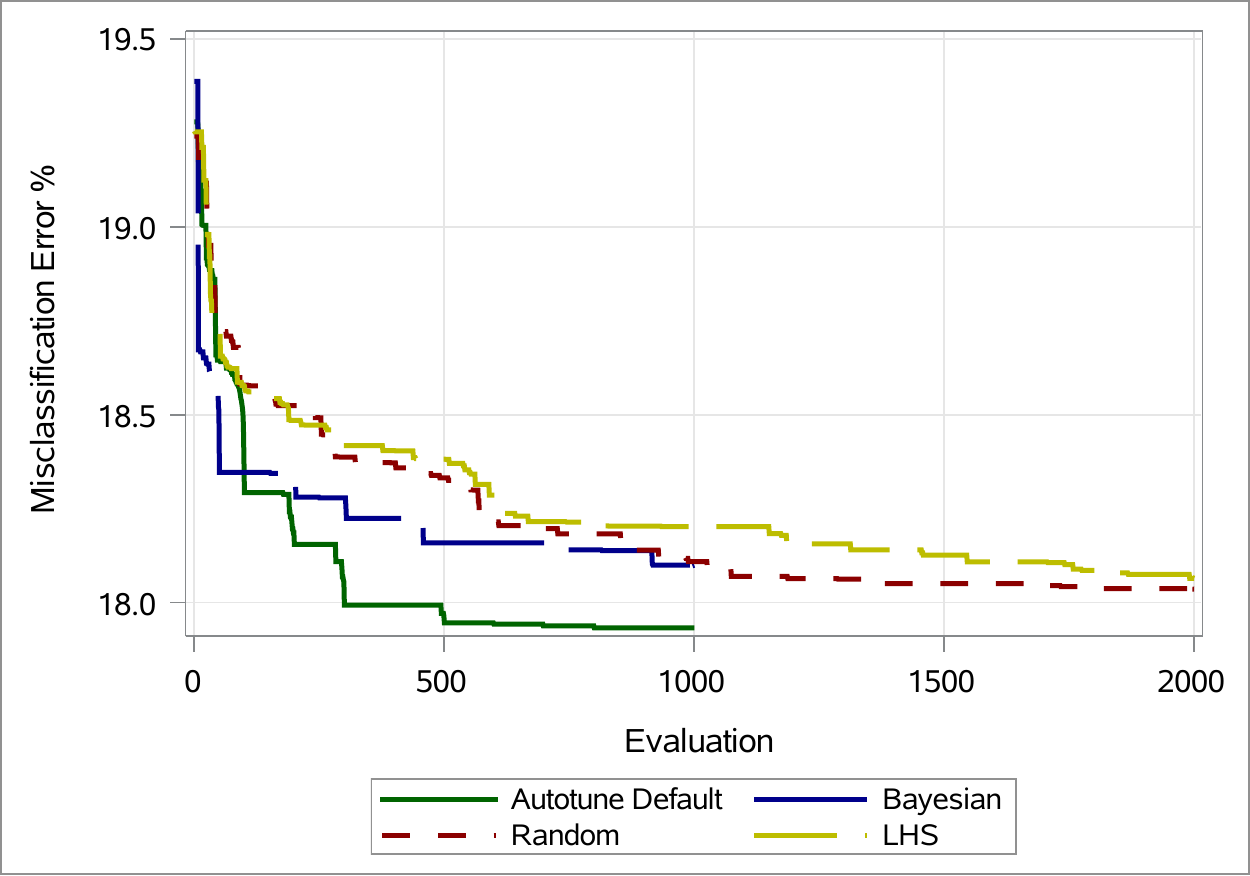}
\caption{The autotune results of  wine quality data set. } 
\label{fig9}
\end{figure}

The wine quality data set is a prepared and extended version of a data set obtained from 
the UCI machine learning repository ~\cite{Lichman}. The data set is a collection of red and white 
variants of the Portuguese ``Vino Verde'' wine ~\cite{Cortez}, with  6,497,000 
observations and 11 features representing 
the physiochemical properties of the wines and a quality rating for each wine.  For the purposes
of this study, the quality ratings were binned such that quality $ \le 6$ was labelled as
``Economy'', and quality $ > 6$ was labeled as ``Premium'', making it a binary classification
problem to predict the new QualityGrp category. In addition, the data set was augmented to 
make it 1000 times larger by synthesizing variations of each observation with random perturbations 
of each attribute value while maintaining the QualityGrp value. 

The compute grid used for the wine data study contains 145 machines:  a controller node and 144 worker nodes.  
Here 4 worker nodes are used for each model training, and a limit of 25 hyperparameter configurations is allowed 
to be evaluated concurrently.
The default Autotune search method is configured with a population size of 101
(resulting in 4 batches of 25 plus the initial/best configuration).  The number of iterations is set to 10,
resulting in up to 1001 hyperparameter configurations evaluated (including the initial configuration). 
 Random search is set to 2000 plus the initial configuration for a 2X random comparison.  Bayesian search is 
 performed with 20 iterations of 50, updating the approximated model after two parallel batches of 25, 
 with a matching maximum number of evaluations of 1000.
 For this study, we use a neural network model and tune the seven hyperparameters as listed in Section \ref{section4}. 
 Each search method is executed 10 times to average random effects. 
For each search method, the tuning time ranges from 2 to 6 hours, and ten repeats of each runs for over 1 day.  
 
Tuning results for the wine data are shown in Figure 9.  The Autotune default search strategy 
converges at a higher rate and to a lower error than the other search methods.  The Bayesian search method beats 
the default search method for the first 150 evaluations, after which its rate of improvement slows.  
It should be noted here that the Gaussian process model is limited in the number of evaluations used to build the 
model; in this case due to the high expense of tuning, with the total of 10 repeats taking over 50 hours, 
the model size is limited to 300 evaluations used in the model.  The Bayesian search method still finds better 
solutions than random search through its 1000 evaluations, after which random search exceeds the capability 
of the limited model used for Bayesian search.  The best solution found by the default search method is better 
than that found by twice as many random search evaluations.


\section{Conclusions}

In this paper, we have presented the hybrid derivative-free optimization framework Autotune for
automated parallel hyperparameter tuning. The system implementation supports multi-level parallelism
where objective evaluations (different model configurations to be trained and validated) 
can be evaluated in parallel across different worker nodes in a grid environment 
while each objective evaluation also uses multiple worker nodes for model training, 
allowing scaling to large data sets and increased training efficiency. 
One lesson learned in applying the system is that the most efficient distributed grid configuration 
for a single model train is usually not the most efficient grid configuration for model tuning. 
More gains are seen from training many models in parallel than making each model train as efficient as possible; 
 careful resource allocation and management of parallel processes is necessary. 
 Furthermore, the framework facilitates concurrent, parallel execution of search methods, 
 sharing of objective evaluations across search methods, easy addition of new search methods, 
 and combining of search methods to create new hybrid strategies, exploiting the strengths of each method.  
 This powerful combination has shown promising numerical results for hyperparameter tuning, 
 where black-box machine learning algorithm complexities include mixed variable types, stochastic and discontinuous objective functions, 
 and the potential for high computational cost.  Combining sampling, local and global search has shown to
 be more robust than applying a single method, and is the main reason why the default search method in Autotune 
 consistently performs better than other search methods. Future work to further enhance Autotune includes 
 improving Autotune's Bayesian search method, handling early stopping of unpromising model configurations 
 more effectively, and supporting multi-objective tuning where trade-offs  between model quality and model complexity can be explored. 


\begin{acks}
  The authors would like to thank the anonymous referees of KDD 2018 for
  their valuable comments and helpful suggestions. 
\end{acks}

\bibliographystyle{ACM-Reference-Format}
\balance 
\bibliography{autotune-bibliography}


\begin{thebibliography}{29}


\ifx \showCODEN    \undefined \def \showCODEN     #1{\unskip}     \fi
\ifx \showDOI      \undefined \def \showDOI       #1{#1}\fi
\ifx \showISBNx    \undefined \def \showISBNx     #1{\unskip}     \fi
\ifx \showISBNxiii \undefined \def \showISBNxiii  #1{\unskip}     \fi
\ifx \showISSN     \undefined \def \showISSN      #1{\unskip}     \fi
\ifx \showLCCN     \undefined \def \showLCCN      #1{\unskip}     \fi
\ifx \shownote     \undefined \def \shownote      #1{#1}          \fi
\ifx \showarticletitle \undefined \def \showarticletitle #1{#1}   \fi
\ifx \showURL      \undefined \def \showURL       {\relax}        \fi
\providecommand\bibfield[2]{#2}
\providecommand\bibinfo[2]{#2}
\providecommand\natexlab[1]{#1}
\providecommand\showeprint[2][]{arXiv:#2}

\bibitem[\protect\citeauthoryear{Bergstra and Bengio}{Bergstra and
  Bengio}{2012}]%
        {BeBe12}
\bibfield{author}{\bibinfo{person}{James Bergstra} {and}
  \bibinfo{person}{Yoshua Bengio}.} \bibinfo{year}{2012}\natexlab{}.
\newblock \showarticletitle{Random Search for Hyper-parameter Optimization}.
\newblock \bibinfo{journal}{\emph{J. Mach. Learn. Res.}}  \bibinfo{volume}{13}
  (\bibinfo{date}{Feb} \bibinfo{year}{2012}), \bibinfo{pages}{281--305}.
\newblock
\showISSN{1532-4435}
\urldef\tempurl%
\url{http://dl.acm.org/citation.cfm?id=2188385.2188395}
\showURL{%
\tempurl}


\bibitem[\protect\citeauthoryear{Cortez, Cerdeira, Almeida, Matos, and
  Reis}{Cortez et~al\mbox{.}}{2009}]%
        {Cortez}
\bibfield{author}{\bibinfo{person}{P. Cortez}, \bibinfo{person}{A. Cerdeira},
  \bibinfo{person}{F. Almeida}, \bibinfo{person}{T. Matos}, {and}
  \bibinfo{person}{J. Reis}.} \bibinfo{year}{2009}\natexlab{}.
\newblock \showarticletitle{Modeling wine preferences by data mining from
  physicochemical properties}.
\newblock \bibinfo{journal}{\emph{In Decision Support Systems}}
  \bibinfo{volume}{47}, \bibinfo{number}{4} (\bibinfo{year}{2009}),
  \bibinfo{pages}{547--553}.
\newblock
\showISSN{0167-9236}


\bibitem[\protect\citeauthoryear{Dewancker, McCourt, Clark, Hayes, Johnson, and
  Ke}{Dewancker et~al\mbox{.}}{2016}]%
        {DeMcClHaJoKe16}
\bibfield{author}{\bibinfo{person}{Ian Dewancker}, \bibinfo{person}{Michael
  McCourt}, \bibinfo{person}{Scott Clark}, \bibinfo{person}{Patrick Hayes},
  \bibinfo{person}{Alexandra Johnson}, {and} \bibinfo{person}{George Ke}.}
  \bibinfo{year}{2016}\natexlab{}.
\newblock \showarticletitle{A Stratified Analysis of Bayesian Optimization
  Methods}.
\newblock \bibinfo{journal}{\emph{CoRR}}  \bibinfo{volume}{abs/1603.09441}
  (\bibinfo{year}{2016}).
\newblock
\showeprint[arxiv]{1603.09441}
\urldef\tempurl%
\url{http://arxiv.org/abs/1603.09441}
\showURL{%
\tempurl}


\bibitem[\protect\citeauthoryear{from SAS Software GitHub~Repository}{from SAS
  Software GitHub~Repository}{2017}]%
        {bankdata}
\bibfield{author}{\bibinfo{person}{Open~Source from SAS Software
  GitHub~Repository}.} \bibinfo{year}{2017}\natexlab{}.
\newblock \bibinfo{title}{Bank Data}.
\newblock
\newblock
\urldef\tempurl%
\url{https://github.com/sassoftware/sas-viya-machine-learning/tree/master/data/bank}
\showURL{%
\tempurl}


\bibitem[\protect\citeauthoryear{Goldberg}{Goldberg}{1989}]%
        {Go89}
\bibfield{author}{\bibinfo{person}{David~E. Goldberg}.}
  \bibinfo{year}{1989}\natexlab{}.
\newblock \bibinfo{booktitle}{\emph{Genetic Algorithms in Search, Optimization
  and Machine Learning} (\bibinfo{edition}{1st} ed.)}.
\newblock \bibinfo{publisher}{Addison-Wesley Longman Publishing Co., Inc.},
  \bibinfo{address}{Boston, MA, USA}.
\newblock
\showISBNx{0201157675}


\bibitem[\protect\citeauthoryear{Gomes, Prud\^{e}ncio, Soares, Rossi, and
  Carvalho}{Gomes et~al\mbox{.}}{2012}]%
        {GoPrSoRoCa12}
\bibfield{author}{\bibinfo{person}{Taciana A.~F. Gomes},
  \bibinfo{person}{Ricardo B.~C. Prud\^{e}ncio}, \bibinfo{person}{Carlos
  Soares}, \bibinfo{person}{Andr{\'e} L.~D. Rossi}, {and}
  \bibinfo{person}{Andr{\'e} Carvalho}.} \bibinfo{year}{2012}\natexlab{}.
\newblock \showarticletitle{Combining Meta-learning and Search Techniques to
  Select Parameters for Support Vector Machines}.
\newblock \bibinfo{journal}{\emph{Neurocomput.}} \bibinfo{volume}{75},
  \bibinfo{number}{1} (\bibinfo{date}{Jan} \bibinfo{year}{2012}),
  \bibinfo{pages}{3--13}.
\newblock
\showISSN{0925-2312}
\urldef\tempurl%
\url{https://doi.org/10.1016/j.neucom.2011.07.005}
\showDOI{\tempurl}


\bibitem[\protect\citeauthoryear{Gray and Fowler}{Gray and Fowler}{2011}]%
        {GrFo11}
\bibfield{author}{\bibinfo{person}{G.~A. Gray} {and} \bibinfo{person}{K.~R.
  Fowler}.} \bibinfo{year}{2011}\natexlab{}.
\newblock \showarticletitle{The Effectiveness of Derivative-Free Hybrid Methods
  for Black-Box Optimization}.
\newblock \bibinfo{journal}{\emph{International Journal of Mathematical
  Modeling and Numerical Optimization}}  \bibinfo{volume}{2}
  (\bibinfo{year}{2011}), \bibinfo{pages}{112--133}.
\newblock


\bibitem[\protect\citeauthoryear{Gray, Fowler, and Griffin}{Gray
  et~al\mbox{.}}{2010}]%
        {GrFoGr10}
\bibfield{author}{\bibinfo{person}{G.~A. Gray}, \bibinfo{person}{K.~R. Fowler},
  {and} \bibinfo{person}{J.~D. Griffin}.} \bibinfo{year}{2010}\natexlab{}.
\newblock \showarticletitle{Hybrid Optimization Schemes for Simulation-Based
  Problems}.
\newblock \bibinfo{journal}{\emph{Procedia Computer Science}}
  \bibinfo{volume}{1} (\bibinfo{year}{2010}), \bibinfo{pages}{1349--1357}.
\newblock


\bibitem[\protect\citeauthoryear{Griffin, Fowler, Gray, and Hemker}{Griffin
  et~al\mbox{.}}{2011}]%
        {GrFoGrHe11}
\bibfield{author}{\bibinfo{person}{J.~D. Griffin}, \bibinfo{person}{K.~R.
  Fowler}, \bibinfo{person}{G.~A. Gray}, {and} \bibinfo{person}{T. Hemker}.}
  \bibinfo{year}{2011}\natexlab{}.
\newblock \showarticletitle{Derivative-Free Optimization via Evolutionary
  Algorithms Guiding Local Search (EAGLS) for MINLP}.
\newblock \bibinfo{journal}{\emph{Pacific Journal of Optimization}}
  \bibinfo{volume}{7} (\bibinfo{year}{2011}), \bibinfo{pages}{425--443}.
\newblock


\bibitem[\protect\citeauthoryear{Griffin and Kolda}{Griffin and Kolda}{2010a}]%
        {GrKo10}
\bibfield{author}{\bibinfo{person}{J.~D. Griffin} {and} \bibinfo{person}{T.~G.
  Kolda}.} \bibinfo{year}{2010}\natexlab{a}.
\newblock \showarticletitle{Asynchronous Parallel Hybrid Optimization Combining
  DIRECT and GSS}.
\newblock \bibinfo{journal}{\emph{Optimization Methods and Software}}
  \bibinfo{volume}{25} (\bibinfo{year}{2010}), \bibinfo{pages}{797--817}.
\newblock


\bibitem[\protect\citeauthoryear{Griffin and Kolda}{Griffin and Kolda}{2010b}]%
        {GrKo10a}
\bibfield{author}{\bibinfo{person}{Joshua~D. Griffin} {and}
  \bibinfo{person}{Tamara~G. Kolda}.} \bibinfo{year}{2010}\natexlab{b}.
\newblock \showarticletitle{Nonlinearly-constrained Optimization Using
  Heuristic Penalty Methods and Asynchronous Parallel Generating Set Search}.
\newblock \bibinfo{journal}{\emph{Applied Mathematics Research eXpress}}
  \bibinfo{volume}{25}, \bibinfo{number}{5} (\bibinfo{date}{October}
  \bibinfo{year}{2010}), \bibinfo{pages}{36--62}.
\newblock
\urldef\tempurl%
\url{https://doi.org/10.1093/amrx/abq003}
\showDOI{\tempurl}


\bibitem[\protect\citeauthoryear{Jones}{Jones}{2001}]%
        {Jo01}
\bibfield{author}{\bibinfo{person}{D.~R. Jones}.}
  \bibinfo{year}{2001}\natexlab{}.
\newblock \showarticletitle{Taxonomy of Global Optimization Methods Based on
  Response Surfaces}.
\newblock \bibinfo{journal}{\emph{Journal of Global Optimization}}
  \bibinfo{volume}{21} (\bibinfo{year}{2001}), \bibinfo{pages}{345--383}.
\newblock


\bibitem[\protect\citeauthoryear{Jones, Perttunen, and Stuckman}{Jones
  et~al\mbox{.}}{1993}]%
        {JoPeSt93}
\bibfield{author}{\bibinfo{person}{D.~R. Jones}, \bibinfo{person}{C.~D.
  Perttunen}, {and} \bibinfo{person}{B.~E. Stuckman}.}
  \bibinfo{year}{1993}\natexlab{}.
\newblock \showarticletitle{Lipschitzian Optimization Without the Lipschitz
  Constant}.
\newblock \bibinfo{journal}{\emph{J. Optim. Theory Appl.}}
  \bibinfo{volume}{79}, \bibinfo{number}{1} (\bibinfo{date}{Oct.}
  \bibinfo{year}{1993}), \bibinfo{pages}{157--181}.
\newblock
\showISSN{0022-3239}
\urldef\tempurl%
\url{https://doi.org/10.1007/BF00941892}
\showDOI{\tempurl}


\bibitem[\protect\citeauthoryear{Konen, Koch, Flasch, Bartz-Beielstein, Friese,
  and Naujoks}{Konen et~al\mbox{.}}{2011}]%
        {KoKoFlBaFrNa11}
\bibfield{author}{\bibinfo{person}{Wolfgang Konen}, \bibinfo{person}{Patrick
  Koch}, \bibinfo{person}{Oliver Flasch}, \bibinfo{person}{Thomas
  Bartz-Beielstein}, \bibinfo{person}{Martina Friese}, {and}
  \bibinfo{person}{Boris Naujoks}.} \bibinfo{year}{2011}\natexlab{}.
\newblock \showarticletitle{Tuned Data Mining: A Benchmark Study on Different
  Tuners}. In \bibinfo{booktitle}{\emph{GECCO '11: Proceedings of the 13th
  Annual Conference on Genetic andEvolutionary Computation}} (2011-01-01),
  \bibfield{editor}{\bibinfo{person}{Natalio Krasnogor}} (Ed.).
  \bibinfo{pages}{1995--2002}.
\newblock


\bibitem[\protect\citeauthoryear{Li, Jamieson, DeSalvo, Rostamizadeh, and
  Talwalkar}{Li et~al\mbox{.}}{2017}]%
        {LiJaDeRoTa17}
\bibfield{author}{\bibinfo{person}{Lisha Li}, \bibinfo{person}{Kevin Jamieson},
  \bibinfo{person}{Giulia DeSalvo}, \bibinfo{person}{Afshin Rostamizadeh},
  {and} \bibinfo{person}{Ameet Talwalkar}.} \bibinfo{year}{2017}\natexlab{}.
\newblock \showarticletitle{Hyperband: Bandit-Based Configuration Evaluation
  for Hyperparameter Optimization}. In \bibinfo{booktitle}{\emph{Proceedings of
  the International Conference on Learning Representations (ICLR)}}.
\newblock


\bibitem[\protect\citeauthoryear{Lichman}{Lichman}{2013}]%
        {Lichman}
\bibfield{author}{\bibinfo{person}{M. Lichman}.}
  \bibinfo{year}{2013}\natexlab{}.
\newblock \showarticletitle{UCI Machine Learning Repository}.
\newblock  (\bibinfo{year}{2013}).
\newblock
\urldef\tempurl%
\url{http://archive.ics.uci.edu/ml}
\showURL{%
\tempurl}


\bibitem[\protect\citeauthoryear{Lorena and Carvalho}{Lorena and
  Carvalho}{2008}]%
        {LoCa08}
\bibfield{author}{\bibinfo{person}{Ana~Carolina Lorena} {and}
  \bibinfo{person}{Andr\'e~C.P.L.F.de Carvalho}.}
  \bibinfo{year}{2008}\natexlab{}.
\newblock \showarticletitle{Evolutionary tuning of SVM parameter values in
  multiclass problems}.
\newblock \bibinfo{journal}{\emph{Neurocomputing}}  \bibinfo{volume}{71}
  (\bibinfo{date}{10} \bibinfo{year}{2008}), \bibinfo{pages}{3326--3334}.
\newblock


\bibitem[\protect\citeauthoryear{McKay}{McKay}{1992}]%
        {Mc92}
\bibfield{author}{\bibinfo{person}{Michael~D. McKay}.}
  \bibinfo{year}{1992}\natexlab{}.
\newblock \showarticletitle{Latin Hypercube Sampling As a Tool in Uncertainty
  Analysis of Computer Models}. In \bibinfo{booktitle}{\emph{Proceedings of the
  24th Conference on Winter Simulation}} \emph{(\bibinfo{series}{WSC '92})}.
  \bibinfo{publisher}{ACM}, \bibinfo{address}{New York, NY, USA},
  \bibinfo{pages}{557--564}.
\newblock
\showISBNx{0-7803-0798-4}
\urldef\tempurl%
\url{https://doi.org/10.1145/167293.167637}
\showDOI{\tempurl}


\bibitem[\protect\citeauthoryear{mldata.org}{mldata.org}{2009}]%
        {mldata}
\bibfield{author}{\bibinfo{person}{mldata.org}.}
  \bibinfo{year}{2009}\natexlab{}.
\newblock \bibinfo{title}{Machine Learning Data Set Repository}.
\newblock
\newblock
\urldef\tempurl%
\url{http://mldata.org/repository/tags/data/IDA_Benchmark_Repository}
\showURL{%
\tempurl}


\bibitem[\protect\citeauthoryear{Nelder and Mead}{Nelder and Mead}{1965}]%
        {NeMe65}
\bibfield{author}{\bibinfo{person}{J.~A. Nelder} {and} \bibinfo{person}{R.
  Mead}.} \bibinfo{year}{1965}\natexlab{}.
\newblock \showarticletitle{A Simplex Method for Function Minimization}.
\newblock \bibinfo{journal}{\emph{Computer Journal}}  \bibinfo{volume}{7}
  (\bibinfo{year}{1965}), \bibinfo{pages}{308--313}.
\newblock


\bibitem[\protect\citeauthoryear{Plantenga}{Plantenga}{2009}]%
        {Pl09}
\bibfield{author}{\bibinfo{person}{T. Plantenga}.}
  \bibinfo{year}{2009}\natexlab{}.
\newblock \bibinfo{booktitle}{\emph{HOPSPACK 2.0 User Manual (v 2.0.2)}}.
\newblock \bibinfo{type}{{T}echnical {R}eport}. \bibinfo{institution}{Sandia
  National Laboratories}.
\newblock


\bibitem[\protect\citeauthoryear{Renukadevi and Thangaraj}{Renukadevi and
  Thangaraj}{2014}]%
        {ReTh14}
\bibfield{author}{\bibinfo{person}{N.T. Renukadevi} {and} \bibinfo{person}{P
  Thangaraj}.} \bibinfo{year}{2014}\natexlab{}.
\newblock \showarticletitle{Performance analysis of optimization techniques for
  medical image retrieval}.
\newblock \bibinfo{journal}{\emph{Journal of Theoretical and Applied
  Information Technology}}  \bibinfo{volume}{59} (\bibinfo{date}{01}
  \bibinfo{year}{2014}), \bibinfo{pages}{390--399}.
\newblock


\bibitem[\protect\citeauthoryear{Sacks, Welch, Mitchell, and Wynn}{Sacks
  et~al\mbox{.}}{1989}]%
        {SaWeMiWy89}
\bibfield{author}{\bibinfo{person}{Jerome Sacks}, \bibinfo{person}{William~J.
  Welch}, \bibinfo{person}{Toby~J. Mitchell}, {and} \bibinfo{person}{Henry~P
  Wynn}.} \bibinfo{year}{1989}\natexlab{}.
\newblock \showarticletitle{Design and Analysis of Computer Experiments}.
\newblock \bibinfo{journal}{\emph{Statist. Sci.}}  \bibinfo{volume}{4}
  (\bibinfo{year}{1989}), \bibinfo{pages}{409--423}.
\newblock
\urldef\tempurl%
\url{https://doi.org/10.1214/ss/1177012413.
  https://projecteuclid.org/euclid.ss/1177012413}
\showDOI{\tempurl}


\bibitem[\protect\citeauthoryear{SAS}{SAS}{2018}]%
        {Viya}
\bibfield{author}{\bibinfo{person}{SAS}.} \bibinfo{year}{2018}\natexlab{}.
\newblock \showarticletitle{SAS\textregistered Viya\texttrademark: Built for
  innovation so you can meet your biggest analytical challenges}.
\newblock  (\bibinfo{year}{2018}).
\newblock
\urldef\tempurl%
\url{https://www.sas.com/content/dam/SAS/en_us/doc/overviewbrochure/sas-viya-108233.pdf}
\showURL{%
\tempurl}


\bibitem[\protect\citeauthoryear{Snoek, Larochelle, and Adams}{Snoek
  et~al\mbox{.}}{2012}]%
        {Snoek}
\bibfield{author}{\bibinfo{person}{Jasper Snoek}, \bibinfo{person}{Hugo
  Larochelle}, {and} \bibinfo{person}{Ryan~P. Adams}.}
  \bibinfo{year}{2012}\natexlab{}.
\newblock \showarticletitle{Practical Bayesian Optimization of Machine Learning
  Algorithms}.
\newblock \bibinfo{journal}{\emph{Advances in Neural Information Processing
  Systems}} (\bibinfo{year}{2012}).
\newblock


\bibitem[\protect\citeauthoryear{Taddy, Lee, Gray, and Griffin}{Taddy
  et~al\mbox{.}}{2009}]%
        {TaGr09}
\bibfield{author}{\bibinfo{person}{M.~A. Taddy}, \bibinfo{person}{H.~K.~H.
  Lee}, \bibinfo{person}{G.~A. Gray}, {and} \bibinfo{person}{J.~D. Griffin}.}
  \bibinfo{year}{2009}\natexlab{}.
\newblock \showarticletitle{Bayesian Guided Pattern Search for Robust Local
  Optimization}.
\newblock \bibinfo{journal}{\emph{Technometrics}}  \bibinfo{volume}{51}
  (\bibinfo{year}{2009}), \bibinfo{pages}{389�401}.
\newblock


\bibitem[\protect\citeauthoryear{Wexler, Haller, and Myneni}{Wexler
  et~al\mbox{.}}{2017}]%
        {WeHaMy17}
\bibfield{author}{\bibinfo{person}{J. Wexler}, \bibinfo{person}{S. Haller},
  {and} \bibinfo{person}{R. Myneni}.} \bibinfo{year}{2017}\natexlab{}.
\newblock \showarticletitle{An Overview of SAS Visual Data Mining and Machine
  Learning on SAS Viya}. In \bibinfo{booktitle}{\emph{SAS Global Forum 2017
  Conference}}. SAS Institute Inc., \bibinfo{address}{Cary, NC}.
\newblock


\bibitem[\protect\citeauthoryear{Wolpert}{Wolpert}{1996}]%
        {Wo96}
\bibfield{author}{\bibinfo{person}{David~H. Wolpert}.}
  \bibinfo{year}{1996}\natexlab{}.
\newblock \showarticletitle{The Lack of a Priori Distinctions Between Learning
  Algorithms}.
\newblock \bibinfo{journal}{\emph{Neural Comput.}} \bibinfo{volume}{8},
  \bibinfo{number}{7} (\bibinfo{date}{Oct.} \bibinfo{year}{1996}),
  \bibinfo{pages}{1341--1390}.
\newblock
\showISSN{0899-7667}
\urldef\tempurl%
\url{https://doi.org/10.1162/neco.1996.8.7.1341}
\showDOI{\tempurl}


\bibitem[\protect\citeauthoryear{Wolpert and Macready}{Wolpert and
  Macready}{1997}]%
        {WoMa97}
\bibfield{author}{\bibinfo{person}{D.~H. Wolpert} {and} \bibinfo{person}{W.~G.
  Macready}.} \bibinfo{year}{1997}\natexlab{}.
\newblock \showarticletitle{No Free Lunch Theorems for Optimization}.
\newblock \bibinfo{journal}{\emph{Trans. Evol. Comp}} \bibinfo{volume}{1},
  \bibinfo{number}{1} (\bibinfo{date}{April} \bibinfo{year}{1997}),
  \bibinfo{pages}{67--82}.
\newblock
\showISSN{1089-778X}
\urldef\tempurl%
\url{https://doi.org/10.1109/4235.585893}
\showDOI{\tempurl}


\end{thebibliography}

\end{document}